# Classification of Alzheimer's Disease Structural MRI Data by Deep Learning Convolutional Neural Networks

## Saman Sarraf [1,2] , Ghassem Tofighi [3]

*samansarraf@ieee.org , gtofighi@ryerson.ca*

[1] Department of Electrical and Computer Engineering, McMaster University, Hamilton, ON, Canada

[2] Rotman Research Institute at Baycrest, University of Toronto, ON, Canada

[3] Electrical and Computer Engineering Department, Ryerson University, Toronto, ON, Canada

# Abstract

Recently, machine learning techniques especially predictive modeling and pattern recognition in biomedical sciences from drug delivery system to medical imaging has become one of the important methods which are assisting researchers to have deeper understanding of entire issue and to solve complex medical problems. Deep learning is a powerful machine learning algorithm in classification while extracting low to high-level features. In this paper, we used convolutional neural network to classify Alzheimer's brain from normal healthy brain. The importance of classifying this kind of medical data is to potentially develop a predict model or system in order to recognize the type disease from normal subjects or to estimate the stage of the disease. Classification of clinical data such as Alzheimer's disease has been always challenging and most problematic part has been always selecting the most discriminative features. Using Convolutional Neural Network (CNN) and the famous architecture LeNet-5, we successfully classified structural MRI data of Alzheimer's subjects from normal controls where the accuracy of test data on trained data reached 98.84%. This experiment suggests us the shift and scale invariant features extracted by CNN followed by deep learning classification is most powerful method to distinguish clinical data from healthy data in fMRI. This approach also enables us to expand our methodology to predict more complicated systems.

# Introduction

## Alzheimer's Disease

Alzheimer's disease is a neurological, irreversible, progressive brain disorder and multifaceted disease that slowly destroys brain cells causing memory and thinking skills loss, and ultimately the ability to carry out the simplest tasks. The cognitive decline caused by this disorder ultimately leads to dementia. For instance, the disease begins with mild deterioration and gets progressively worse in a neurodegenerative type of dementia. Diagnosing Alzheimer's disease requires very careful medical assessments such as patients' history, Mini Mental State Examination (MMSE) and physical and neurobiological exam. Structural imaging based on magnetic resonance is an integral part of the clinical assessment of patients with suspected Alzheimer dementia. Atrophy of medial temporal structures is now considered to be a valid diagnostic marker at the mild cognitive impairment stage. Structural imaging is also included in diagnostic criteria for the most prevalent non-Alzheimer dementias, reflecting its value in differential diagnosis. In addition, rates of whole-brain and hippocampal atrophy are sensitive markers of neurodegeneration, and are increasingly used as outcome measures in trials of potentially disease-modifying therapies [1]. Clinical diagnostic criteria are currently based on the clinical examination and neuropsychological assessment, with the identification of dementia and then of the Alzheimer's phenotype [2]. Development of an assistant tool or algorithm to classify structural MRI data and more importantly to recognize brain disorder data from healthy subjects has been always clinicians 'interests. Any machine learning algorithm which is able to classify Alzheimer's disease assists scientists and clinicians to diagnose this brain disorder. In this work, the convolutional neural network (CNN) which is one of the Deep Learning Network architecture is utilized in order to classify the Alzheimer's brains and healthy brains and to produce a trained and predictive model.

## Deep Learning

Hierarchical or structured deep learning is a modern branch of machine learning that was inspired by human brain. This technique has been developed based on complicated algorithms that model high-level features and extract those abstractions from data by using similar neural network architecture but much complicated. The neuroscientists discovered the "neocortex" which is a part of the cerebral cortex concerned with sight and hearing in mammals, process sensory signals by propagating them through a complex hierarchy over time. That was the main motivation to develop the deep machine learning focusing on computational models for information representation that exhibit similar characteristics to that of the neocortex [3] [4] [5]. Convolutional Neural Networks which are inspired by human visual system are similar to classic neural networks. This architecture has been particularly designed based on the explicit assumption that raw data are two-dimensional (images) that enables us to encode certain properties and also to reduce the amount of hyper parameters. The CNN topology utilizes spatial relationships to reduce the number of parameters which must be learned and thus improves upon general feed-forward back propagation training. Equation 1 shows how Error is calculated in the back propagation step where E is error function, y is the $i^{th}$, $j^{th}$ neuron, x is the input, l represent layer numbers w is filter weight with a and b indices, N is the number of neurons in a given layer and m is the filter size.

$$\frac{\partial E}{\partial y_{ij}^{l-1}} = \sum_{a=0}^{m-1} \sum_{b=0}^{m-1} \frac{\partial E}{\partial x_{(i-a)(j-b)}^{l}} \frac{\partial x_{(i-a)(j-b)}^{l}}{\partial y_{ij}^{l-1}} = \sum_{a=0}^{m-1} \sum_{b=0}^{m-1} \frac{\partial E}{\partial x_{(i-a)(j-b)}^{l}} w_{ab}$$

**Equation 1: This shows how the error of backpropagation is calculated.**

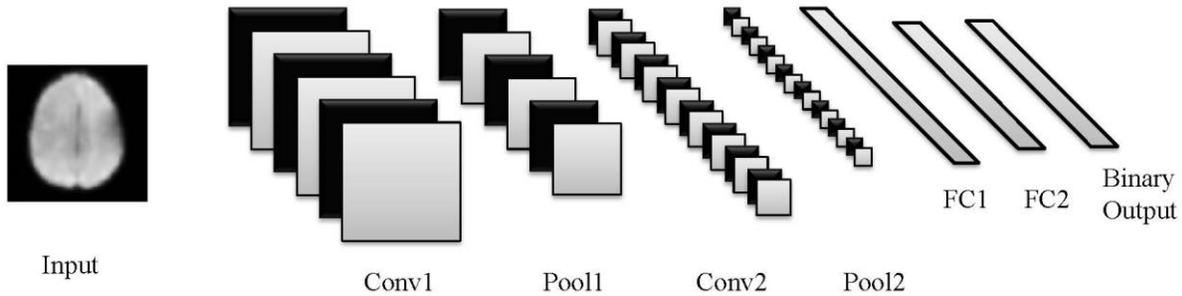

**Figure 1: The architecture of LeNet-5 includes two convolutional layers, two pooling layers, two fully connected layers and a softmax layer which outputs a binary decision in this classification problem.**

Deeper CNN architectures were developed to recognize numerous objects from high volume data such as AlexNet (ImageNet) [6], ZF Net [7], GoogleNet [8], VGGNet [9] and ResNet [10]. GoogLeNet developed by Szegedy et al. [8] is a successful network which is broadly used for object recognition and classification.

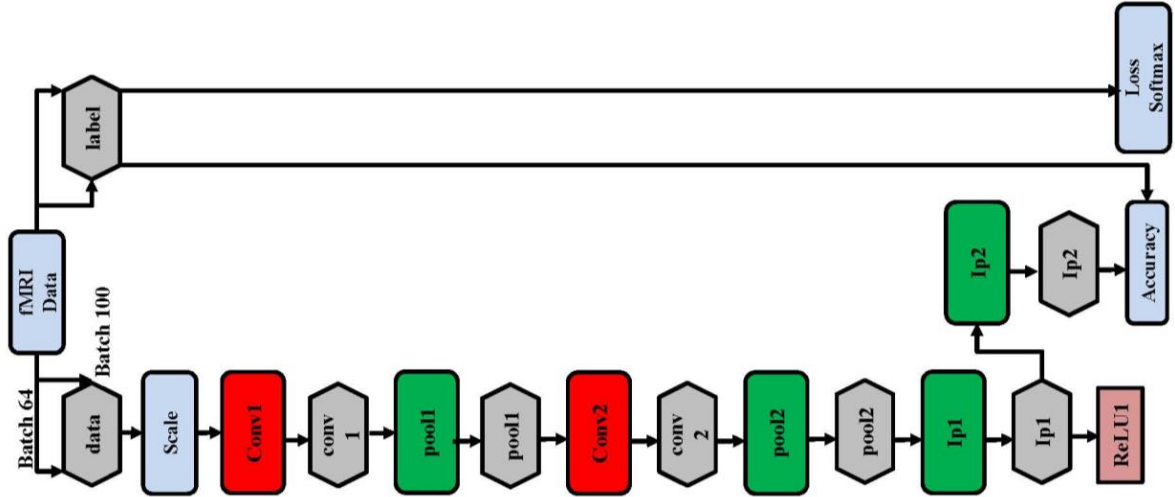

Figure 2: Deep convolutional neural network architecture was proposed by Szegedy et al. [8] codenamed Inception that achieves the new state of the art for classification and detection in the ImageNet Large-Scale Visual Recognition Challenge 2014 (ILSVRC14). The main hallmark of this architecture is the improved utilization of the computing resources inside the network. One particular incarnation used in our submission for ILSVRC14 is called GoogLeNet, a 22 layers deep network, the quality of which is assessed in the context of classification and detection.

## Data Acquisition and Preprocessing

### ADNI Database

The structural MRI dataset included 302 subjects whose structural magnetic resonance imaging data (MRI) were acquired (age group > 75). This group included 211 Alzheimer's Disease patients and 91 healthy control. Certain subjects were scanned in different time points whose imaging data were separately considered in this work. **Error! Reference source not found.** is presented the demographic information of both subsets including mini–mental state examination (MMSE) scores.

Table 1: In ADNI database, the subjects older than 75 years were selected who had structural MRI data. The demographic data are shown in the table below.

| Modality | Total Subj. | Group | Subj. | Female | Mean of Age | SD | Male | Mean of Age | SD | MMSE ± SD |
|---|---|---|---|---|---|---|---|---|---|---|
| MRI | 302 | Alzheimer | 211 | 85 | 80.98 | 21.6 | 126 | 81.27 | 16.66 | 23.07 ± 2.06 |
| | | Control | 91 | 43 | 79.37 | 12.52 | 48 | 80.81 | 19.51 | 28.81 ± 1.35 |

MRI data acquisition was performed according to the ADNI acquisition protocol [11]. Scanning was performed on different 3 Tesla scanners: General Electric (GE) Healthcare, Philips Medical Systems, and Siemens Medical Solutions based on identical scanning parameters. Anatomical scans were acquired with a 3D MPRAGE sequence (TR=2s, TE=2.63 ms, FOV=25.6 cm, 256 x 256 matrix, 160 slices of 1mm thickness). Functional scans were acquired by using an EPI sequence (150 volumes, TR=2 s, TE=30 ms, flip angle=70, FOV=20 cm, 64 x 64 matrix, 30 axial slices of 5mm thickness without gap).

## Structural MRI data preprocessing

The raw data of structural MRI scans for both AD and NC group were provided in NII format in ADNI database. First, all non-brain tissues were removed from images using Brain Extraction Tool FSL-BET [12] by optimizing the fractional intensity threshold and reducing image bias and residual neck voxels. Then, the study-specific grey matter template was created using FSL-VBM library and protocol found at http://fsl.fmrib.ox.ac.uk/fsl/fslwiki/FSLVBM [13]. In this step all brain-extracted images were segmented to grey matter (GM), white matter (WM) and Cerebrospinal fluid (CSF). GM images were selected and registered to GM ICBM-152 standard template using linear affine transformation. The registered images were concatenated and averaged then flipped along the x-axis and the two mirror images then re-averaged to obtain a first-pass, study-specific affine GM template. Second, the GM images are re-registered to this affine GM template using non-linear registration, concatenated into a 4D image called, averaged, flipped along the x-axis. Both mirror images were then averaged to create the final symmetric, study-specific "non-linear" GM template at 2x2x2 $mm^3$ resolution in standard space. After that, all concatenated and averaged 3D GM images (one 3D image per subject) were concatenated into a stack (4D image = 3D images across subjects) called "Structural MRI". Also, the FSL-VBM protocol introduced a compensation or modulation for the contraction/enlargement due to the non-linear component of the transformation where each voxel of each registered grey matter image is multiplied by the Jacobian of the warp field. The modulated 4D image was then smoothed by a range of Gaussian kernels; sigma = 2, 3, 4 mm (standard sigma values in the field of MRI data analysis) approximately resulted in FWHM 0f 4.6, 7 and 9.3 mm. The various spatial smoothing kernels enabled us to explore if the classification accuracy would improve. The MRI preprocessing module was applied to AD and NC data and produced two sets of four 4D images which were called Structural MRI 0 - fully preprocessed without smoothing - and three fully preprocessed and smoothed datasets called Structural MRI 2, 3, 4 used in next steps for classification.

## Structural MRI pipeline

The preprocessed MRI data including were loaded into memory and by a similar approach to fMRI pipeline converted from NII to PNG format using Nibabel and OpenCV which created two groups (AD and NC) × four preprocessed datasets (MRI 0,2,3,4). Also, the last 10 slices of subject as well as slices with zero mean pixels were removed from the data. This step produced a total 62335 images where 52507 belonged to AD and the remaining 9828 belonged to NC group per dataset. The data were then converted to LMDB format while resized to 28x28 pixels. The adopted LeNet model was set for 30 epochs and initiated for Stochastic Gradient Descent with gamma = 0.1, momentum = 0.9, base learning rate = 0.01, weight_decay = 0.0005 and Step learning rate policy dropping the learning rate in steps by a factor of gamma every stepsize iterations. Next, the model was trained and tested by 75% and 25% of the data for four different datasets. The training and testing process was repeated five times on Amazon AWS Linux G2.8xlarge to assure the robustness of network and achieved accuracy. The average of accuracies was obtained for each experiment separately shown in **Error! Reference source not found.**. The results demonstrated that a high level of accuracy was achieved in all the experiments where the highest accuracy of 98.79% was achieved for the structural MRI dataset that was spatially smoothed by sigma = 3 mm. In the second run, the adopted GoogleNet model was selected for this binary classification. In this experiment, the preprocessed datasets were converted to LMDB format where resized to 256x256. The model was adjusted for 30 epochs using Stochastic Gradient Descent with gamma = 0.1, momentum = 0.9, base learning rate = 0.01, weight_decay = 0.0005 and Step learning rate policy. The GoogleNet model resulted in a higher level of accuracy than LeNet where the highest overall accuracy of 98.8431% was achieved for MRI 3 (smoothed by sigma = 3mm). However, the accuracy of unsmoothed dataset (MRI 0) reached 0.845043 which lower than the similar experiment with LeNet model which can show the negative effect of interpolation on unsmoothed data which can strengthen the concept of spatial smoothing in MRI data analysis. In practice, most classification questions deal with imbalanced data which refers to a classification problem where the data are not represented equally where the ratio of data may exceed 4 to 1 in binary classification. In the MR analyses performed in this study, the ratio of AD to NC used for training the CNN classifier was around 5 to 1. To validate the accuracies of models developed, a new set of training and testing was performed by randomly selecting and decreasing the number of AD images to 10722 for training while the same number of images 9828 for NC group was used. In the balanced data experiment, the adopted LeNet model was adjusted for 30 epochs using the same parameters mentioned above and trained for four MRI datasets. In table 2, the new results are shown in with labels starting with B. prefix (Balanced). The highest accuracy obtained from the balanced data experiment only decreased around 1% (B. Structural MRI 3 = 0.9781) compared to the same datasets in the original training. This comparison demonstrated that the new results were highly correlated to the previous ones showing even decreasing the data ratio from 5:1 to 1:1 had no impact on the classification accuracy which validated the robustness of the trained models in the original MRI classification.

Table 2: The accuracies of "testing" data for three AD classifications were obtained where an accuracy of 98.84% was achieved for Adopted GoogleNet.

| Dataset | Architecture | Averaged Accuracy |
|---|---|---|
| Structural MRI 0 | Adopted LeNet | 0.97446 |
| Structural MRI 2 | | 0.98566 |
| Structural MRI 3 | | 0.9879 |
| Structural MRI 4 | | 0.98672 |
| Structural MRI 0 | Adopted GoogleNet | 0.845043 |
| Structural MRI 2 | | 0.98452 |
| Structural MRI 3 | | 0.988431 |
| Structural MRI 4 | | 0.987758 |
| B. Structural MRI 0 | Adopted LeNet | 0.9572 |
| B. Structural MRI 2 | | 0.975 |
| B. Structural MRI 3 | | 0.9781 |
| B. Structural MRI 4 | | 0.9746 |

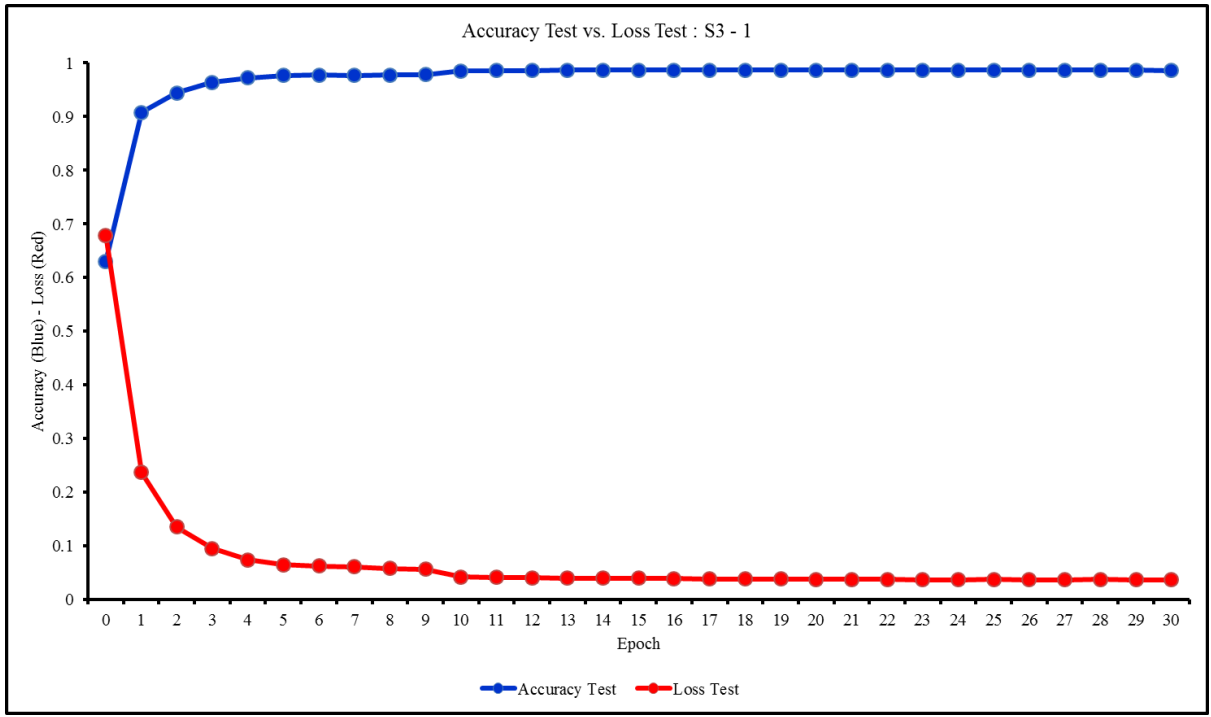

**Figure 3: The accuracy and loss of testing data for 30 epochs using the adopted LeNet architecture are shown in this figure for the structural MRI dataset that are processed and smoothed by sigma = 3 mm.**

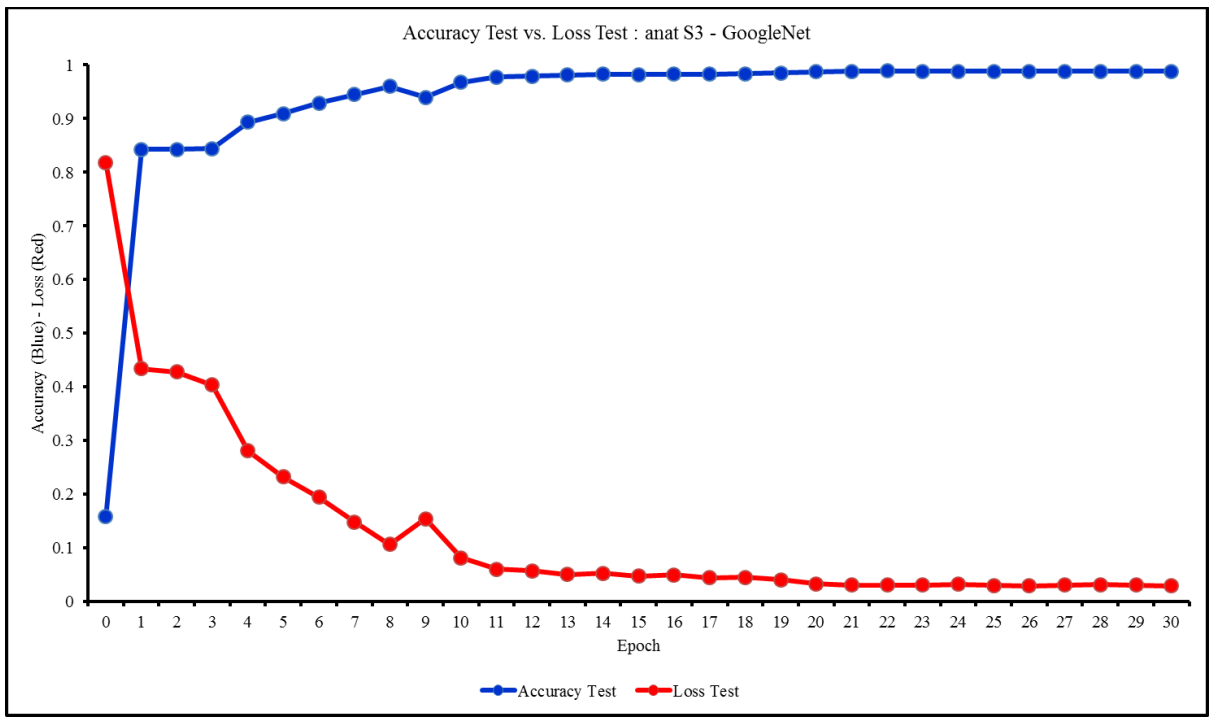

**Figure 4: The accuracy and loss of testing data for 30 epochs using the adopted GoogleNet architecture are shown.**

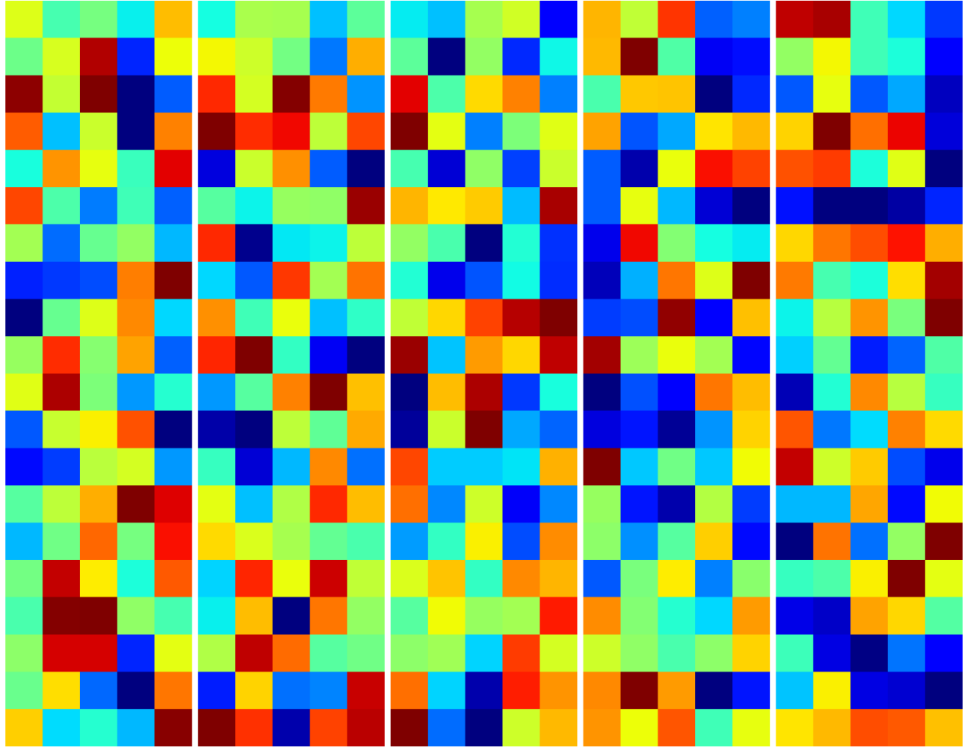

Figure 5: The filters (20, 5, 5) of the first Convolution layer for LeNet architecture are displayed.

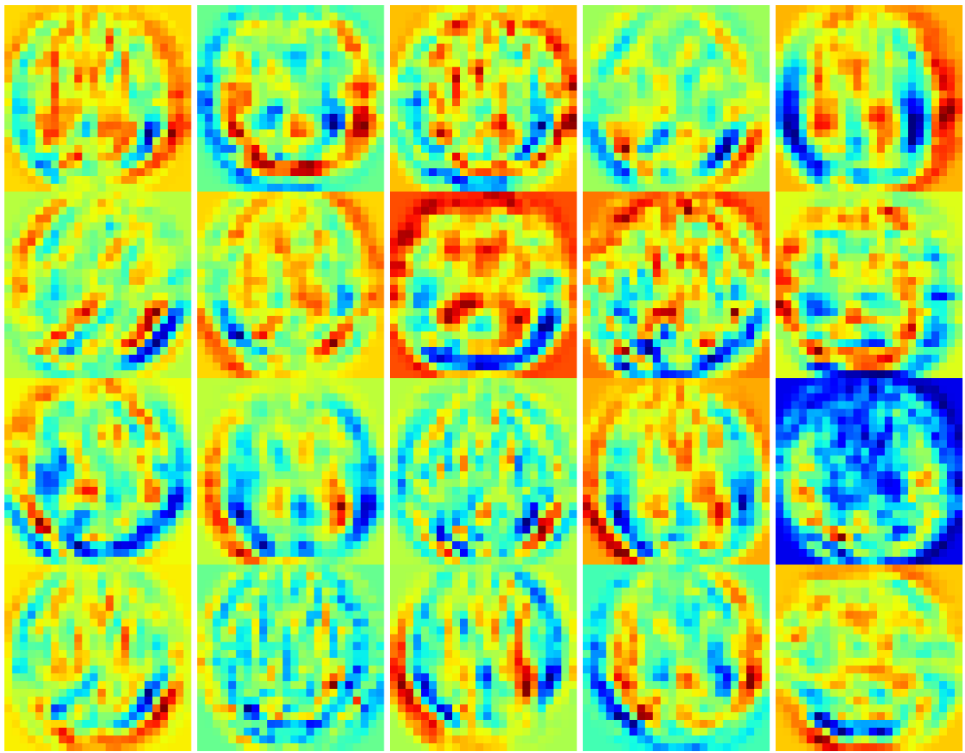

Figure 6: The features of first Convolutional layer in LeNet architecture for a given AD subject are shown.

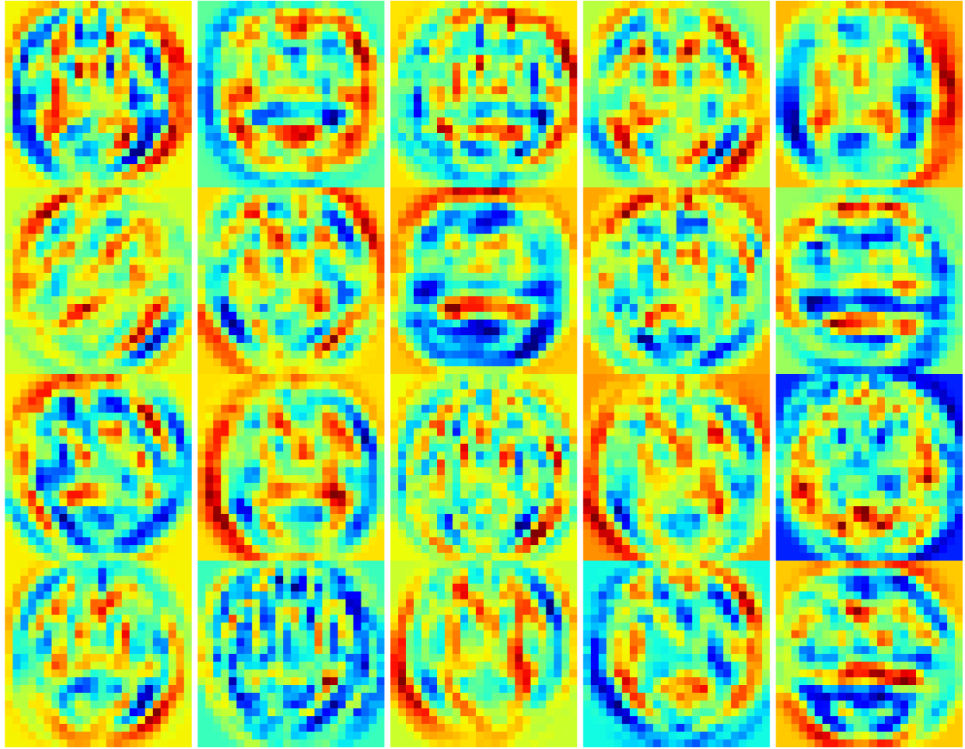

Figure 7: The features of first Convolutional layer for a given NC subject are shown. (LeNet architecture)

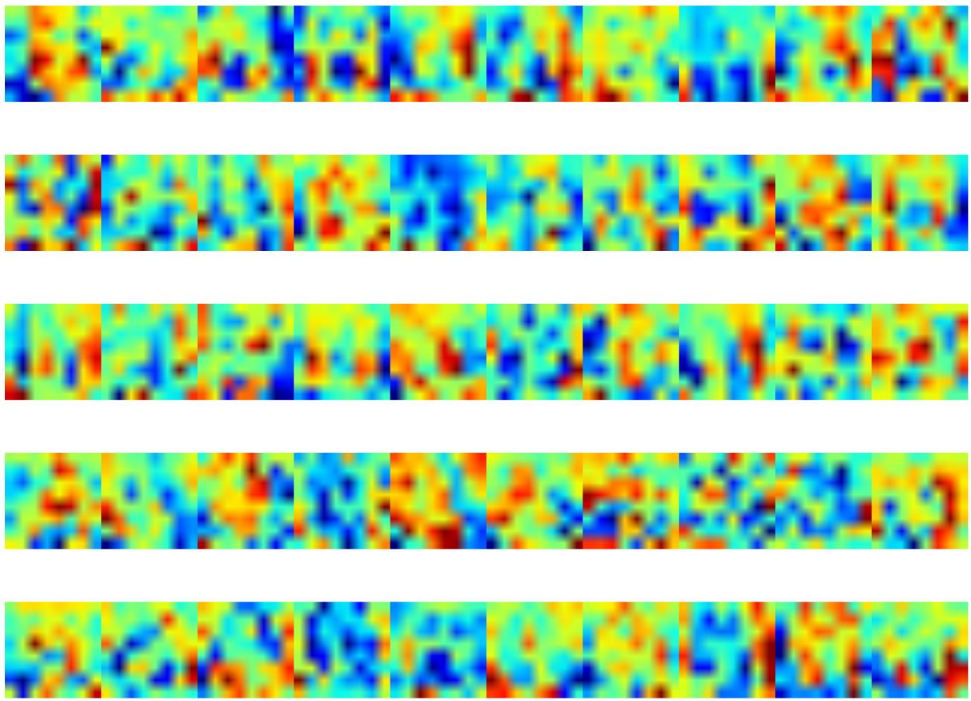

Figure 8: The features (50, 8, 8) of the second Convolution layer and a given AD brain for LeNet architecture are displayed.

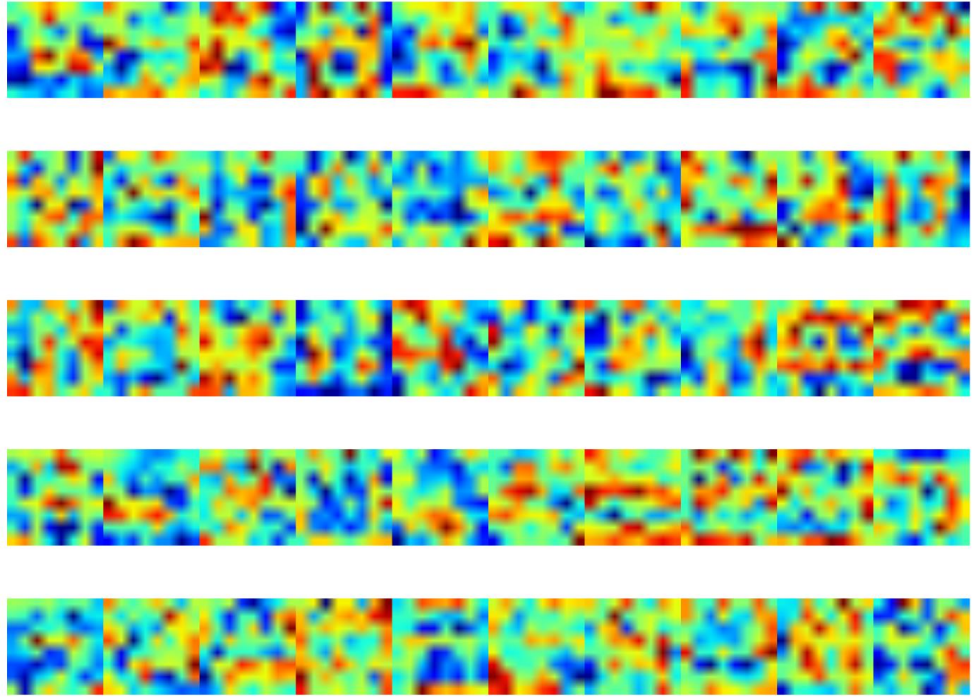

**Figure 9: The features (50, 8, 8) of the second Convolution layer and a given NC brain for LeNet architecture are displayed.**

## Conclusion

In this work, using structural MRI data and also two CNN architectures: LeNet and GoogleNet we successfully predicted AD from NC brains where an accuracy of 98.84% was achieved in the best-case scenario. Advances in deep learning enabled us to use a very deep CNN structure adopted for our binary classification method. The shift and scale invariant features extracted from different layers of CNN architecture resulted in a highly accurate trained model. Furthermore, extensive and unique preprocessing strategies utilized in this work improved the quality of the data fed into LeNet and GoogleNet which ultimately positively impacted the classifier performance. This paper demonstrated high potential of using deep learning architectures opening new avenues for medical diagnostic imaging especially from brain disorders.